\documentclass[letterpaper, 10 pt, conference]{ieeeconf}  

\IEEEoverridecommandlockouts                              

\usepackage{amsmath,amssymb,amsfonts}
\usepackage{algorithmic}
\usepackage{graphicx}
\usepackage{textcomp}
\usepackage{xcolor}
\usepackage{verbatim}
\usepackage{comment}
\usepackage[super]{nth}

\usepackage{lipsum}
\usepackage{cuted}
\usepackage{capt-of}
\usepackage[utf8]{inputenc}
\usepackage{dirtytalk}
\usepackage[normalem]{ulem}
\usepackage{soul}
\usepackage{algorithm2e}  
\usepackage{hyperref}

\usepackage{tabularx}
\usepackage{color, colortbl}
\definecolor{Gray}{gray}{0.8}

\title{\LARGE \bf
PufferBot: Actuated Expandable Structures for Aerial Robots
}

\author{
Hooman Hedayati$^{1}$,
Ryo Suzuki$^{1, 2}$,
Daniel Leithinger$^{3}$,
and Daniel Szafir$^{3}$
\thanks{
$^{1}$Department of Computer Science, University of Colorado Boulder.
{\tt\small \href{mailto:hooman.hedayati@colorado.edu}{\color{blue}hooman.hedayati@colorado.edu}}}
\thanks{
$^{2}$Department of Computer Science, University of Calgary.
{\tt\small \href{mailto:ryo.suzuki@ucalgary.ca}{\color{blue}ryo.suzuki@ucalgary.ca}}}
\thanks{
$^{3}$Department of Computer Science and ATLAS Institute, University of Colorado Boulder.
{\tt\small \href{mailto:daniel.leithinger@colorado.edu}{\color{blue}daniel.leithinger},\href{mailto:daniel.szafir@colorado.edu}{\color{blue}daniel.szafir}
{\color{blue}{@colorado.edu}}}}
}

\begin{document}

\maketitle

\begin{strip}\centering
 \vspace{-3.5cm}
\begin{center}
\begin{tabular}{ l } 
\includegraphics[width=\textwidth]{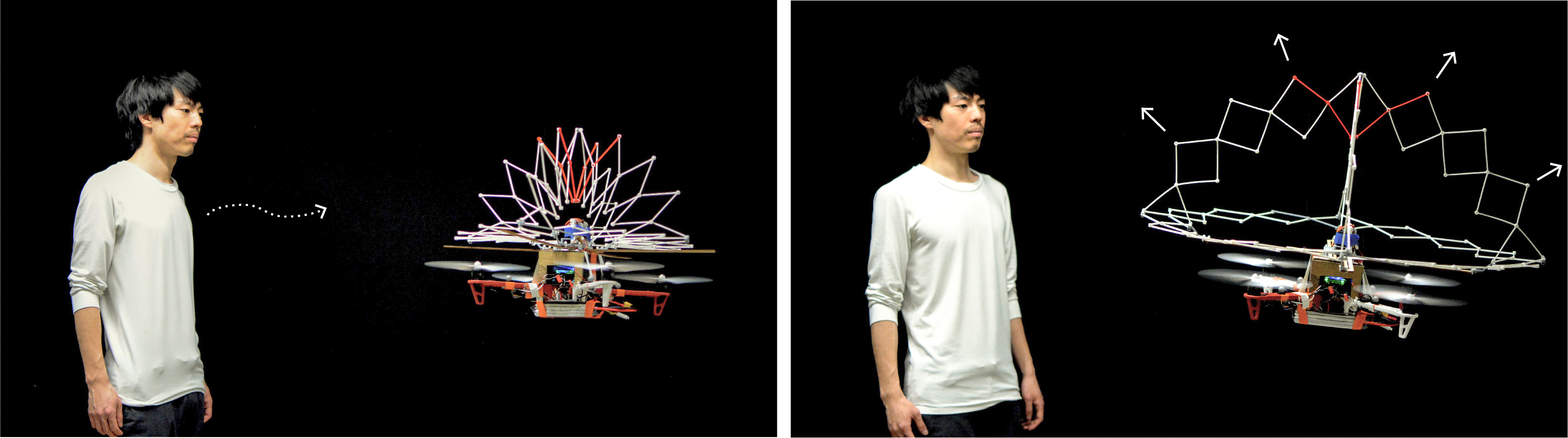} \\ 
PufferBot is an aerial robot augmented with an actuated, expandable structure that may expand to protect the robot or \\ collocated humans in the event of a collision. \\ 

\end{tabular}
\end{center}
\vspace{-.5cm}

\label{fig:teaser}
\end{strip}

\thispagestyle{empty}
\pagestyle{empty}

\begin{abstract}

We present PufferBot, an aerial robot with an expandable structure that may expand to protect a drone's propellers when the robot is close to obstacles or collocated humans. PufferBot is made of a custom 3D-printed expandable scissor structure, which utilizes a one degree of freedom actuator with rack and pinion mechanism. We propose four designs for the expandable structure, each with unique characterizations for different situations.
Finally, we present three motivating scenarios in which PufferBot may extend the utility of existing static propeller guard structures. The supplementary video can be found at:  \href{https://youtu.be/XtPepCxWcBg}{\color{blue}https://youtu.be/XtPepCxWcBg} 

\end{abstract}

\section{INTRODUCTION}
Aerial robots are increasingly used in a wide variety of applications, such as search and rescue, journalism, structural inspection, and environmental data collection. When used indoors, aerial robots have traditionally been isolated from humans through cages or operated in an entirely separated space, but they are increasingly entering into environments with collocated humans (e.g., construction sites). In such situations, there is an increasing demand to reduce the danger and unpredictability of robots, as well as increase safety for nearby people. At a high level, aerial robots introduce a major safety consideration beyond those present in traditional ground robots as many types of hardware and/or software failures will lead to complications in  maintaining the robot's altitude, with the robot subsequently falling onto the ground or crashing into obstacles. Moreover, aerial robot propellers are often quite fragile, with slight damage to the propellers leading to instabilities and/or errors in how the robot executes a plan from a flight controller. Any crashes resulting from falls or erroneous flight paths may damage the robot and create dangers for nearby humans. 

To address issues surrounding propeller damage/safety, many aerial robots use propeller guards---fixed structures that may prevent the propellers from hitting an obstacle or person in the event of a collision. However, many guards do not fully cover the robot's propellers (for instance, only providing cover for the horizontal size of a propeller), leaving other parts of the propellers (e.g., the top) exposed and vulnerable to damage. On the other hand, guards that do provide full coverage surrounding the propellers, such as in the Zero Zero Robotics HoverCam~\cite{zerozerorobotics} and the Flyability GimBall~\cite{flyability}, significantly increase the size and rigidity of the robot, potentially making the robot less maneuverable. This can pose a problem if the robot operates in narrow spaces (e.g., search and rescue in a collapsed building), as the robot cannot navigate tight spaces and can become stuck between obstacles. Finally, such systems only provide a static \textit{buffer zone} indicating appropriate interaction distances between the robot and any nearby humans.

In this paper, we introduce PufferBot, the concept of an expandable aerial robot that can dynamically change its shape to reduce damage in the event of collisions with collocated humans and/or the environment. PufferBot consists of an aerial robot with a mounted expandable structure that can be actuated to expand in order to reduce the collision damage or create an enlarged buffer zone surrounding the robot. The PufferBot concept is inspired by both natural designs (e.g., pufferfish) and mechanical systems (e.g., vehicle airbags). When in danger, a pufferfish (Tetraodontidae) inflates its body by taking water or air into portions of its digestive tract to increase its size. Similarly, vehicle airbag systems also inflate to protect humans when crashes occur. 

By taking an inspiration from such metaphors, we propose an expandable structure for an aerial robot that may reduce the risk of crashing and protect the robot's propellers when the robot is in danger of falling on the ground, crashing into an object, or navigating cluttered spaces. One advantage of our system is that the expandable structure can dynamically change its shape in order to reduce the overall size in the non-expanded state, making it easier for the robot to navigate in narrow spaces and avoid unnecessary contact with the surrounding environment. In addition, such expandable structures may open up a new design element for future work examining user interaction (e.g., using robot expansion/contraction as a communicative mechanism, similar to~\cite{suzuki2019shapebots,suzuki2020roomshift, takei2011kinereels}).

In this paper, we first describe related work in robotic safety and expandable structures. We then explain our design and implementation of the PufferBot system and present applications of PufferBot. Finally, we discuss PufferBot's limitations and our planned future work.

\begin{figure}[!t]
\centering
\includegraphics[width=\linewidth]{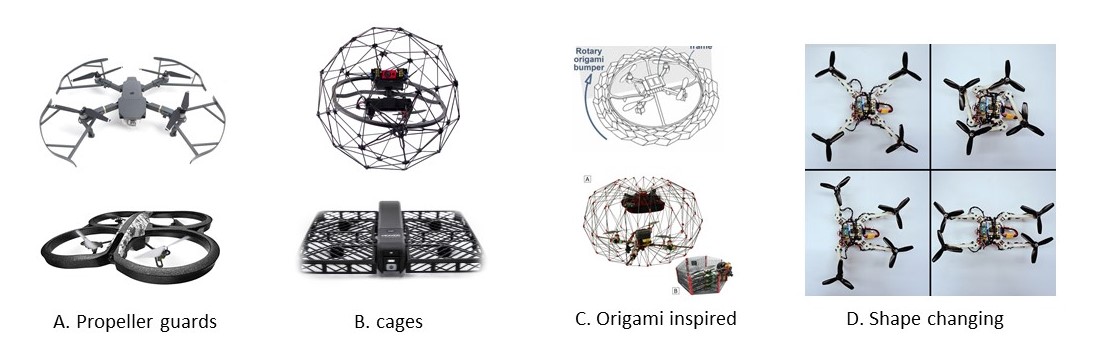}
\caption{The existing approaches.}
\label{fig:relatedWork}
\end{figure}

\section{Related work}
Our work is informed by prior research in the field of physical human-robot interaction (pHRI), which has examined problem of integrating robots in human-populated environments. The literature on pHRI has identified several methods in which safety can be ensured, including safety through control, motion planning, prediction, and a consideration of human psychological factors. 
Relevant to our particular concern of physical safety, past research typically focuses on one or more components of the collision management pipeline, including 
 \textit{pre-collision}, which usually deals with a combination of human detection and prediction \cite{kuhn2007fast,ragaglia2018trajectory}, maintaining safe separation distance \cite{svarny2019safe}, or avoiding and predicting undesired collisions via motion planning and control \cite{flacco2012depth,rossi2015pre,khatib1986real,madarash2004enhancing};
 \textit{collision detection, isolation and identification}, to understand the severity of a collision should one occurs outside of the prescribed robot operation (\cite{flacco2016residual,de2012integrated,golz2015using});
 and \textit{post-collision} and post-impact phase (\cite{haddadin2017robot,golz2015using,parusel2011modular}); see \cite{lasota2017survey} for an overall survey of such methods and \cite{haddadin2016physical} for a survey focused on pHRI.

\begin{figure}[!b]
\includegraphics[width=\linewidth]{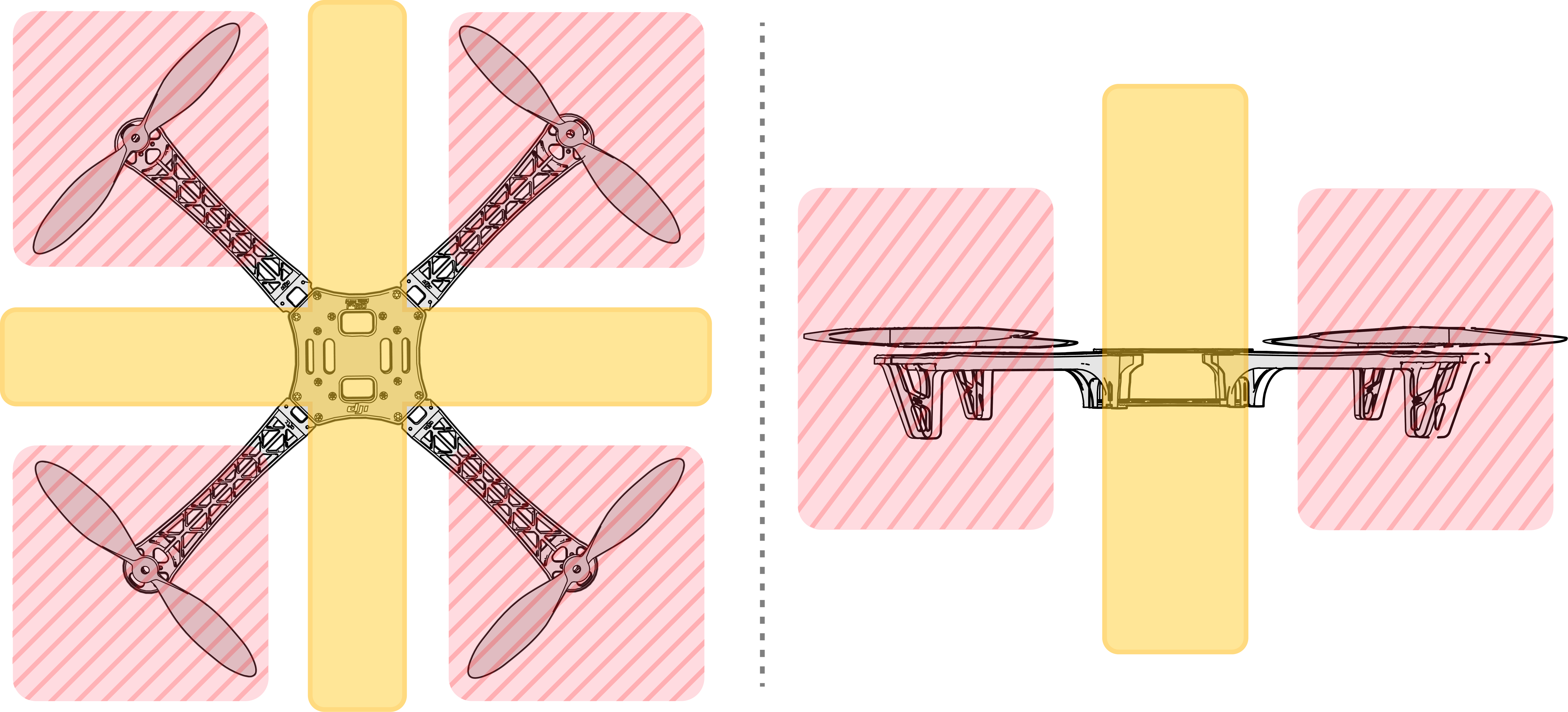}
\caption{Yellow area shows the two geometric planes which can be used for an actuator of expandable structure}
\label{fig:spacelimitation}
\end{figure}

\begin{figure*}[!t]
\includegraphics[width=\linewidth]{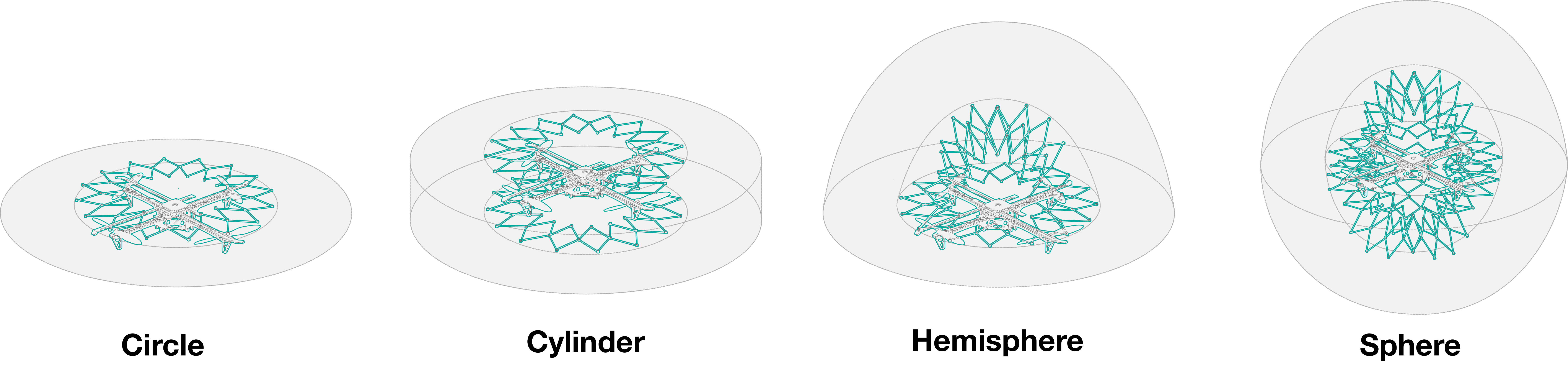}
\caption{The four designs that can be implemented.
}
\label{fig:limitation}
\end{figure*}


Within this general framework, drone/UAV researchers have mostly tried to improve aerial robot perception systems to better detect humans \cite{aguilar2017pedestrian} and avoid collisions in the first place (i.e., pre-collision management) \cite{lioulemes2014safety, padhy2018monocular}. Many of these efforts do not directly target safety but indirectly result in safer human-robot interactions. As an alternative approach, other research on aerial robots has focused on how to communicate planned robot flight trajectories to users to address the \textit{motion inference} problem whereby interactions involving robots and collocated humans may be unsafe (or perceived as unsafe) if humans are unable to understand how, when, and where a robot will move (and thus cannot plan their own movements accordingly). Within this line of research, prior work has explored how robot  motion \cite{szafir2015communicating}, lights \cite{szafir2014communication}, and augmented reality visualizations \cite{hedayati2018improving,walker2018communicating,walker2019robot,gadre2019end,szafir2019mediating} may make the robot's actions/motion more predictable.

A third method for improving aerial robot safety is through enhancing the hardware design of the robot itself. For example, Pounds et al. \cite{pounds2018safety} developed a passively spinning hoop as a mechanical interference sensor for detecting obstacles and triggering an electrodynamic braking system. This system results in instant landing (i.e., falling to the ground) which may not always be ideal although the notion of a braking system that does not injure humans is relevant to our motivation. To date, most efforts in hardware design have been made by hobbyist/consumer aerial robot companies that produce propeller guards (Fig. \ref{fig:relatedWork}A), an optional robot component that is often mounted on each arm of the aerial robot. Such guards, which may be made of soft (e.g., foam) or rigid (e.g., plastic) materials may provide the robot with a minimal form of impact reduction while reducing any damage that may be caused by spinning propellers to humans or objects the robot crashes into. 
While many guards only protect the propellers along a single axis (most often, the horizontal axis), certain commercial systems use protective cages (Fig. \ref{fig:relatedWork}B) that cover the whole drone and completely isolated the propellers from the surrounding environment. Two particular examples of aerial robot cages are Zero Zero Robotics HoverCamera~\cite{zerozerorobotics} and and Flyability GimBall~\cite{flyability} (Fig. \ref{fig:relatedWork}B).

While simple propeller guards and full cages can provide some protection for the robot propellers (and likewise protect humans and the environment from being damaged), 
such static systems are limited in providing only minimal compliance, may increase the size and rigidity of the robot frame (potentially reducing maneuverability), and do not necessarily provide nearby humans with many cues regarding safety (potentially leading to issues of over or undertrust in the system). 

Recent work has begun to examine new aspects of aerial robot design that may address some of these limitations. For example, our work draws inspiration from research that explores dynamic propeller shells \cite{salaan2019development}, foldable drone frames \cite{falanga2018foldable}, and origami-inspired mechanisms for aerial robots \cite{kornatowski2017origami, shu2019quadrotor} (Fig. \ref{fig:relatedWork}C,D). 
Sareh et al.'s \cite{sareh2018rotorigami} work developing an origami-based circlular protection guard for drones is particularly relevant to our work, although this system is still a static structure compared to our PufferBot, which focuses on dynamic, actuated expandable structures. 



\section{Design}
In this section, we describe our design choices and rationale for the PufferBot. Our design process focused on two potentially competing objectives: (1) safety of the robot, collocated humans, and environment, and (2) robot maneuverability. Safety and maneuverability can often be a trade-off; for example, mounting a cage on a drone's propellers may improve safety but increases the robot's size and weight meaning that the drone may not be able to navigate as confined spaces.

To balance these objectives, we explored  expandable structures,  special structures that can change size or volume in dynamic ways. Expandable structures have been used in masts, arches, plane spatial structures, and cylindrical and spherical bar structures \cite{escrig1993geometry}. One advantageous property of expandable structures is that it is typically easy to change overall shape of the structure by applying force in a single direction, which may reduce the complexity in attempting to actuate such structures. Expandable structures can have different shapes, such as foldable circular structures or foldable ring structures~\cite{you1997foldable}.

For aerial robots, the propeller number, position, and configuration imposes a design constraint for an expandable structure. For example, for a common quadcopter configuration, the space that can be used for actuation of an expandable structure is shown by the yellow areas in Fig. \ref{fig:spacelimitation}.
Based on this design constraint, we propose an expandable structure with a customizable, modular ring design. The ring design is constructed from connected scissor units that make a circle shape. The scissor units consists of two angled arms that are connected to each other at their intermediate points by a revolvable joint~\cite{maden2011review}. The ring design can be easily extended to the other designs, such as a hemisphere by adding two half rings or a full sphere by adding two full rings (all without needing to alter the underlying actuation mechanism).
Below, we describe four different geometries made possible by this customizable ring design: circle, cylinder, hemisphere, and sphere (Fig \ref{fig:spacelimitation}):

{\bf Circle}: This design is made by a single ring and is similar to existing propeller guards on the market. The advantage of the circle over static guards is that it can expand and contract as needed. 
This is the most lightweight design, with a weight of $1 \times Mass_{ring}$. A disadvantage of this design is that it cannot protect the propellers in all directions. If the robot hits an obstacle from above or below, there is potential for damage. 

{\bf Cylinder}: This design is made by two parallel rings connected to each other and mounted on the aerial robot. This is most similar to the static HoverCamera 
cage when the structure is expanded. This Cylinder still may not protect the robot fully, but provides more coverage than the Circle design. A disadvantage of this design is that the rings directly above and below the propellers may change airflow, potentially requiring the flight controller be re-tuned. 
The weight of the Cylinder is $2 \times Mass_{ring}$.

{\bf Hemisphere}: This design is made by a full ring and two half rings. The half rings are perpendicular to each other and to the full ring. As a whole, they make a half sphere. The weight of this structure is the same as the Cylinder ($2 \times Mass_{ring}$). The half sphere protects the upper half of the aerial robot, which is more fragile since the lower part of the robot may be protected by the landing equipment. The Hemisphere design helps protect the robot from certain types of vertical obstacles beyond those addressed by the Circle and Cylinder (e.g., a light hanging from the ceiling). Moreover, it does not interfere with the airflow on the top of the propellers.

{\bf Sphere}: The last design is a full sphere made by three full rings, in which the two additional rings are perpendicular to another. This design is the heaviest of our designs, $3 \times Mass_{ring}$, which can reduce the efficiency of the robot and total flight time. The advantage of the sphere is that it provides the most robust protective structure that fully surrounds the robot (similar to the Flyability GimBall). In addition, as the Sphere design presents a similar appearance regardless of observer angle, it may support distance estimation during line-of-sight teleoperation. 


We can also change the angle and the length of the two bars in scissor units. By modifying these two parameters we can change the radius and smoothness of the ring. The radius of the ring has a linear correlation with the length of each individual part in scissor unit. Through design iteration, we determined that a minimum 15 cm gap was a safe distance between propellers and the safety guard. The length of the arms were selected to satisfy this constraint. The angle in the arm is correlated with the number of scissor unit used in each ring. The more scissor units used, the smoother the ring is. More scissor units make the structure more robust, but increases the weight of the structure and thus may decrease efficiency.  After creating several prototype designs, we found that 16 scissor units provided a good trade-off between the weight and the smoothness of the ring and the stiffness of the structure when encountering obstacles.

\section{Implementation}
\begin{figure}[!htb]
\includegraphics[width=\linewidth]{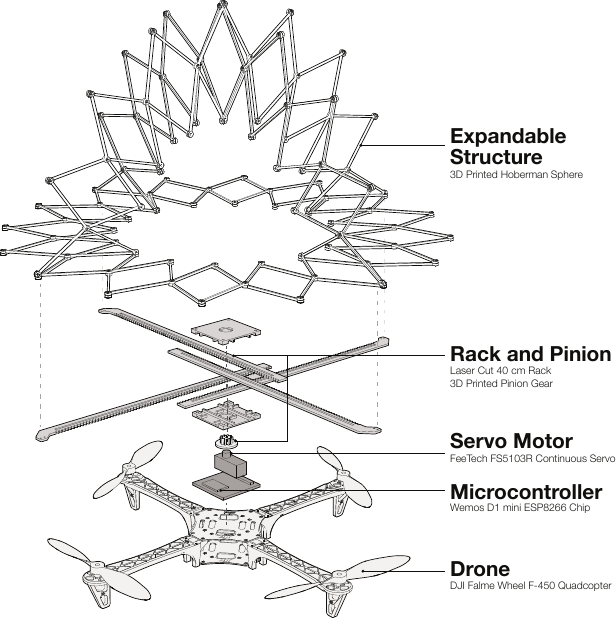}
\caption{PufferBot parts}
\label{fig:sphereStructure}
\end{figure}

In this section, we explain the implementation of the system and specification for each component. The four components of the PufferBot are (1) an aerial robot, (2) a 3D printed expandable structure, (3) a actuation mechanism, and (4) a controller (Fig. \ref{fig:sphereStructure}). We explain each component in detail.

\subsection{Aerial Robot}
In this work, We used a DJI Flame Wheel F450 frame for our aerial robot. The base frame weight is 282 g. After mounting additional components (motors, battery, flight controller, etc.), the weight of the aerial robot accumulates to 1.2 Kg. Based on the specification document, the robot is capable of lifting up to 1.6 Kg of payload. which is enough for our expandable structure (in our setting, the weight of expandable structure and the actuator combined is 1.6 Kg). The diagonal length of the robot (motor to motor) is 45 cm. We used $ 4 \time 4.5$ inch propellers (11.43 cm), which make the total length of the aerial robot 70cm. We used a 4S  Lithium-ion Polymer (LiPo) battery as the power source, which gives the robot a flight time of  approximately 18 minutes. We built a plate on top of the aerial robot which gives us enough surface to mount and secure the expandable structure and actuator. The plate also allows us to avoid direct contact with the inertial measurement unit (IMU) and onboard sensors in the flight controller. 

\subsection{Expandable Structure}
\begin{figure}[!ht]
\includegraphics[width=\linewidth]{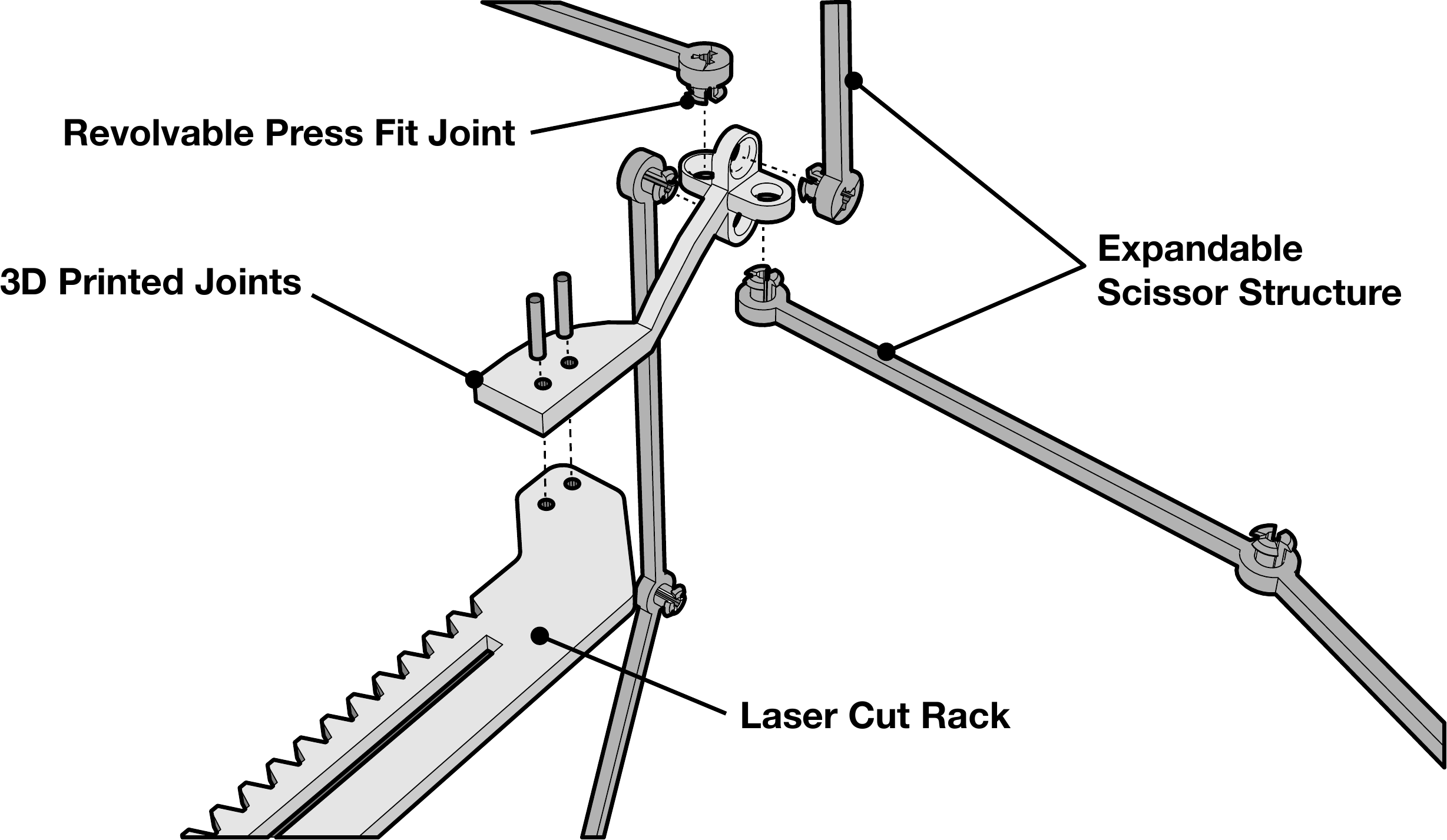}
    \caption{The assembly of the PufferBot.}
\label{fig:assembly}
\end{figure}

The expandable structure is made of three parts: (1) actuator joints, (2) regular joints, and (3) scissor units. 
74 pieces were used in total to make the expandable structure. Revolvable press fit joints are used for secure but rotatable connections (Fig \ref{fig:assembly}).

\subsection{Actuation}
We actuate the expandable structure with a one degree of freedom actuation mechanism based on rack and pinion.
The pinion gear located in the center rotates the four individual rack planes at the same time, so that the actuated racks can evenly apply the expansion force in four different directions at the same rate. The actuator joint attached to the end of the rack can expand and collapse the expandable structure by pushing and pulling the connected points.
All the actuator parts are 3D printed with PLA. We laser cut the racks with 3mm plywood. We used plywood after testing with different materials (e.g., acrylic) and found that plywood was the most robust in terms of holding its shape against bending forces over time. We used a FeeTech FS5103R as a servo motor, controlled by a Wemos D1 mini ESP8266 micro-controller.

\begin{figure}[!ht]
\includegraphics[width=\linewidth]{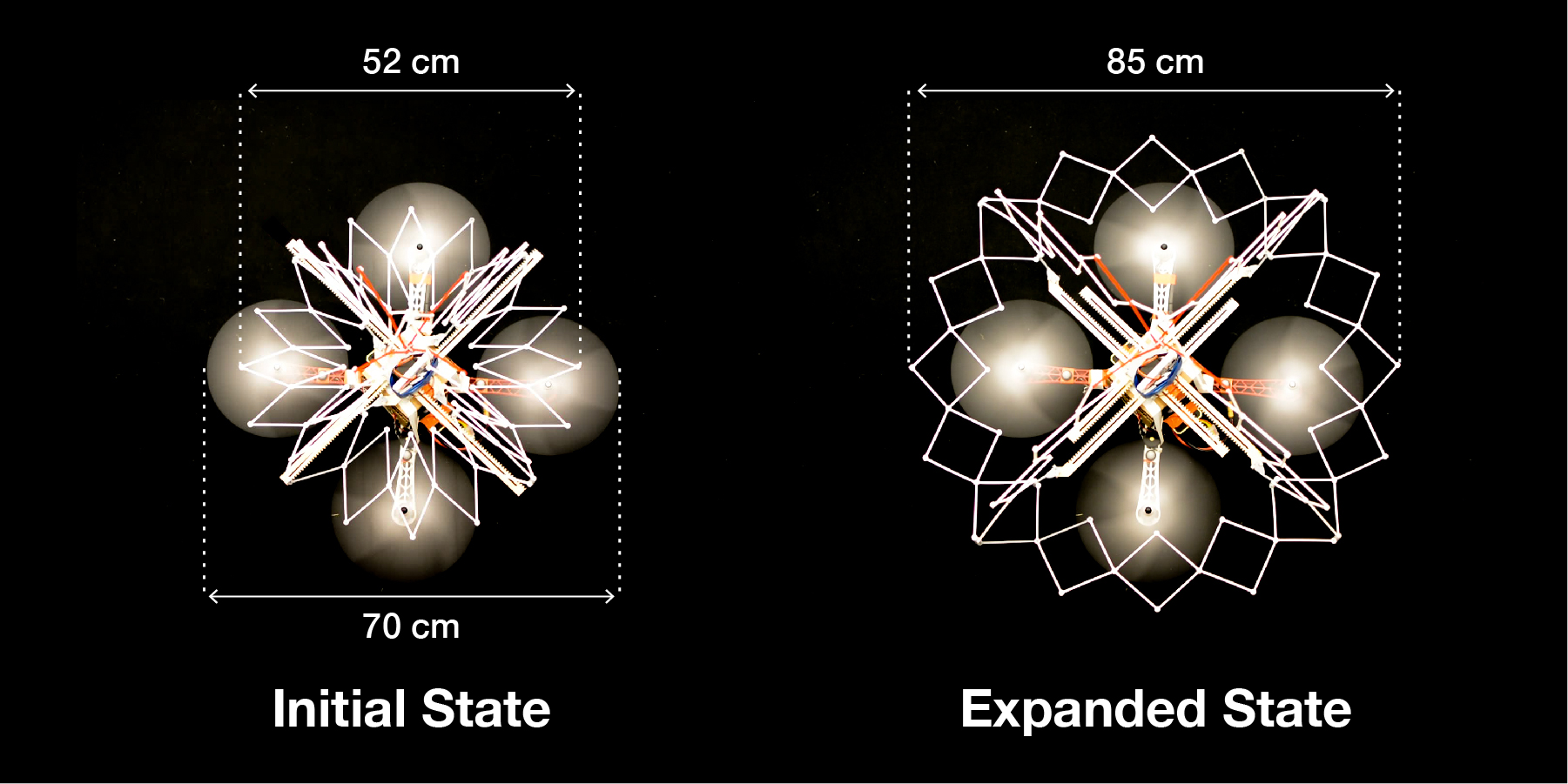}
    \caption{The drone need 70cm clearance, the structure size can vary from 52cm to 85cm}
\label{fig:expanded}
\end{figure}

\subsection{Control}
There are two components that need to be controlled in the PufferBot: the aerial robot and the expandable structure actuator. Both can be controlled autonomously by a central computer or manually by a teleoperator. To implement autonomous control, we developed a linear PID controller that controls the position and altitude of the aerial robot. 
As the aerial robot we used lacks on-board sensing capabilities sufficient for accurate localization, our PID controller currently relies on a Vicon motion capture system with 200Hz motion tracking cameras embedded in the environment to track the physical robot. There is a trajectory planner built on top of the PID controller so the robot can traverse the space as planned. 
The local computer that runs the Vicon system also continuously detects nearby obstacles, and based on this information, it wirelessly communicates with the actuator's microcontroller to programmatically control the size of the expandable structure.
In the manual mode,  a teleoperator is in charge of controlling the robot as well as the expandable structure.  

\subsection{Performance}
The Pufferfish weights 600 grams and can expand or collapse in 6 seconds (we tested the expanding structure 50 times and the number reported is the mean of the trials). PufferBot can handle 6-9 N of force: 6 N against the parts furthest from the actuation racks and up to 9 N applied to the links directly connected to the racks.

\section{Use Cases}
In general, we envision that actuated expandable structures, such as our PufferBot design, will be helpful for robots operating in cluttered spaces and/or in situations where humans are collocated and potentially collaborating with the robot. In this section, we detail several scenarios in which we anticipate PufferBot will be particularly useful.

\begin{figure}[!htb]
\includegraphics[width=\linewidth]{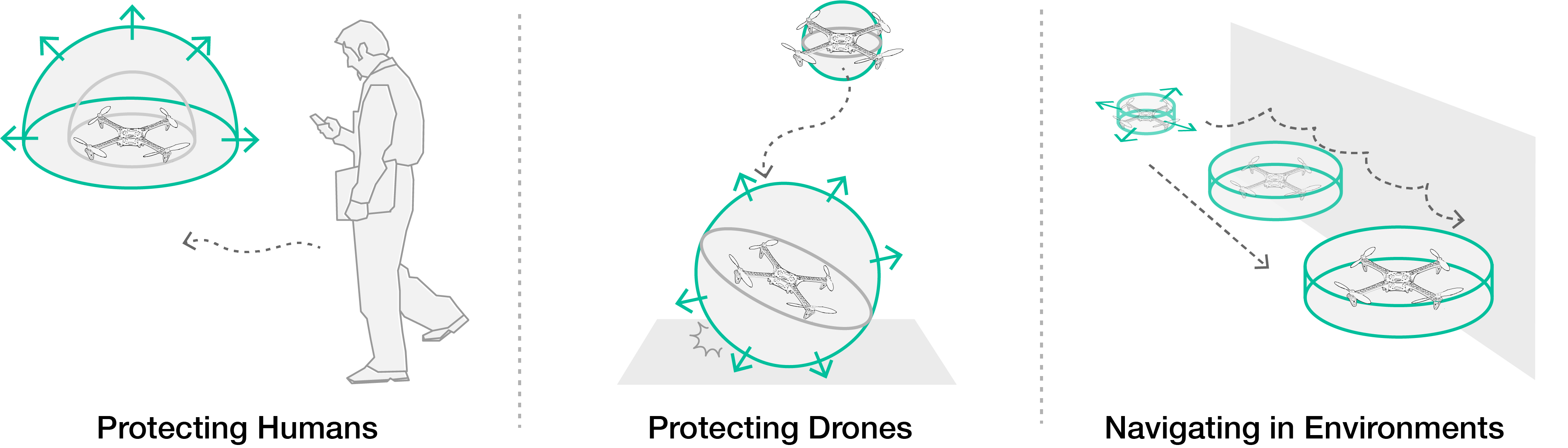}
\caption{We envision PufferBot augmenting aerial robots across a range of scenarios, including helping to alert and protect collocated humans, avoid damage to the robotic platform itself, and support collision-tolerant navigation.}
\label{fig:usecases}
\end{figure}

\vspace{-5pt}
\subsection{Protecting Humans}
Aerial robots and humans are increasingly occupying shared spaces, whether through collaboration in the workplace or users partaking in leisure activities (e.g., a hobbyist piloting an aerial robot in their neighbourhood). While human safety is critical in these situations, 
developing autonomous/semi-autonomous systems capable of mitigating all risk of collisions remains an open problem, with some risks arising from humans themselves (e.g., a wandering human who is not paying attention or one who is attempting to test the robot's behaviors might incite a collision with the robot). In these scenarios, PufferBot may reduce the risk of human injury in contacting the robot's spinning propellers, disperse the force of the robot during a collision over a wider surface area, and provide a compliance mechanism that helps mitigate impact force. Moreover, PufferBot may support pre-collision management by communicating a customizable buffer zone that indicates safe proxemic interaction zones with the robot, potentially expanding to increase the buffer zone as conditions change (e.g., due to the ratio of humans in the environment vs environment size, inferences regarding human knowledge of the robot, etc.). 
In autonomous systems, the expansion capability may be activated in response to sensor readings or as part of a fail-safe (potentially improving the autonomous landing/falling procedure that many systems currently implement when faults arise). For a teleoperator, an emergency button can be implemented on the controller that immediately activates PufferBot, like an airbag.

\subsection{Protecting the Drone}
In recent years, there has been an increasing trend in utilizing aerial robots for inspecting bridges, powerlines, pipelines, and other infrastructure elements. For these tasks, aerial robots must operate close to the target of interest, increasing the chance of the robot hitting obstacles due to operator error, loss of power, and/or unexpected elements of weather like gusts of wind. In these situations, PufferBot may protect the robot from dangerous objects in the environment. In the worst case, PufferBot may reduce damage to the robot when in free-fall by expanding to leverage the scissor structure, which acts like a spring (Fig. \ref{fig:usecases}).

\subsection{Sensing and Navigating Complex Environments}
When drones navigate in complex indoor environments, they usually rely on some form of perception (e.g., SLAM) to avoid colliding with walls and obstacles. However, many conditions may impair perception algorithms, including smoke, glare, dirty camera lenses, etc. We propose to use the expanding scissor structure as a guiding mechanism in collision-tolerant navigation, similar to the whiskers of a cat, the use of white canes by people who are blind or visually impaired, or the bumper bar of wheeled robots like Roomba. When the drone is in a complex environment, it may expand in order to locate obstacles by bumping into them. We can further mount sensors into the structure to enhance this navigation and even provide haptic feedback to a teleoperator to indicate when the presence of nearby obstacles.

\section{CONCLUSIONS}
In this paper we present the design and implementation of PufferBot, an aerial robot with an actuated, expandable structure that serves to protect the robot, collocated humans, and the operating environment. Our modular design supports four different configurations that make trade-offs between the protection offered and the resulting robot maneuverability. We detail the construction of the Hemisphere design in particular to demonstrate an end-to-end proof of concept of our system. Finally, we present several  use cases in which we envision PufferBot providing utility.

\addtolength{\textheight}{-12cm}   




\section*{ACKNOWLEDGMENT}
This work was supported by an Early Career Faculty grant from NASA’s Space Technology Research Grants Program under award NNX16AR58G and by the Nakajima Foundation scholarship and the Strategic Information and Communications R\&D Promotion Programme (SCOPE) of the Ministry of Internal Affairs and Communications of Japan. We thank Vinitha Gadiraju for her help in our research. 

\bibliographystyle{IEEEtran}
\bibliography{main}

\end{document}